\documentclass[10pt,twocolumn,letterpaper]{article}

\usepackage{cvpr}
\usepackage{times}
\usepackage{epsfig}
\usepackage{graphicx}
\usepackage{amsmath}
\usepackage{amssymb}

\usepackage{microtype}
\usepackage{dblfloatfix}
\usepackage{booktabs}
\usepackage{multirow}
\usepackage{listings}
\usepackage{enumitem}

\usepackage[pagebackref=true,breaklinks=true,letterpaper=true,colorlinks,bookmarks=false]{hyperref}
\usepackage{cleveref}
\crefformat{footnote}{#2\footnotemark[#1]#3}

\cvprfinalcopy 

\def\cvprPaperID{****} 

\DeclareMathOperator{\GR}   {GR}
\DeclareMathOperator{\splus}{softplus}
\DeclareMathOperator{\relu} {ReLU}
\DeclareMathOperator{\bigo} {\mathcal{O}}
\newcommand{\LG}{\mathcal{L}_G}
\newcommand{\LD}{\mathcal{L}_D}
\newcommand{\tG}{\theta_G}
\newcommand{\tD}{\theta_D}
\newcommand{\pd}{\partial}
\allowdisplaybreaks

\begin{document}

\title{FusedProp: Towards Efficient Training of Generative Adversarial Networks}

\author{
Zachary Polizzi\thanks{Equal contribution.}\\
South Park Commons\\
{\tt\small zplizzi@gmail.com}
\and
Chuan-Yung Tsai\footnotemark[1]\\
South Park Commons\\
{\tt\small cytsai@gmail.com}
}

\maketitle

\begin{abstract}
Generative adversarial networks (GANs) are capable of generating strikingly realistic samples but state-of-the-art GANs can be extremely computationally expensive to train.
In this paper, we propose the fused propagation (FusedProp) algorithm which can be used to efficiently train the discriminator and the generator of common GANs simultaneously using only one forward and one backward propagation.
We show that FusedProp achieves 1.49 times the training speed compared to the conventional training of GANs, although further studies are required to improve its stability.
By reporting our preliminary results and open-sourcing our implementation, we hope to accelerate future research on the training of GANs.
\end{abstract}

\begin{figure*}
\begin{center}
\includegraphics[width=0.7\textwidth]{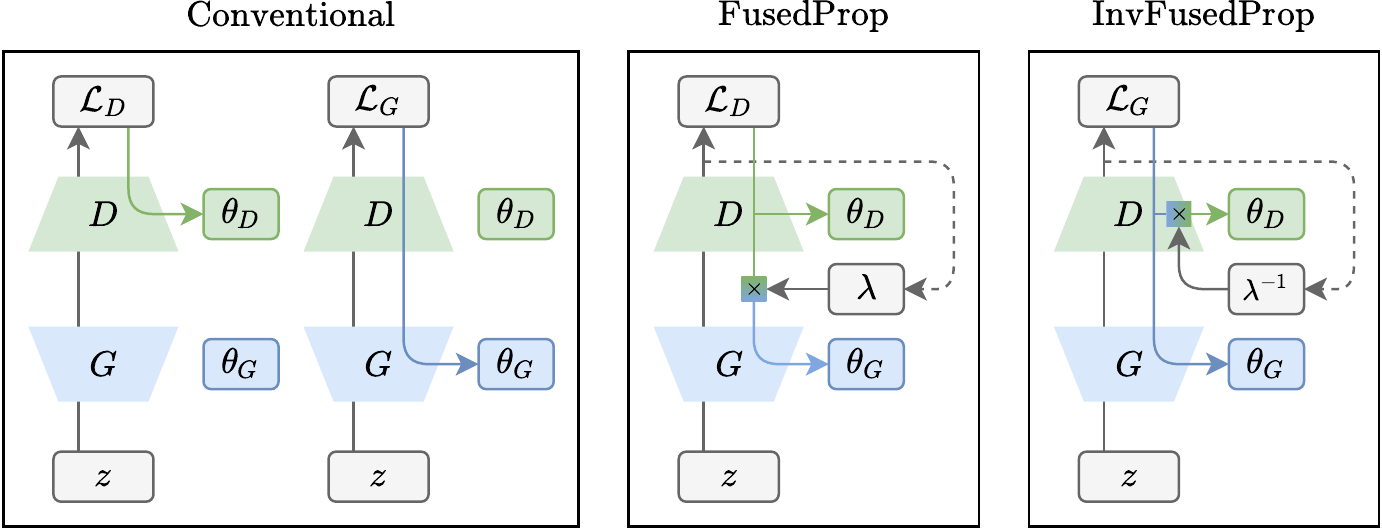}
\end{center}
\caption{Conventional \vs fused propagation (FusedProp) and inverted fused propagation (InvFusedProp) training of GANs, where gray arrow indicates forward propagation, colored arrow indicates backward propagation and dashed arrow indicates conditional dependency. $\LD^R$ is omitted for simplicity.}
\label{fig:fusedprop}
\end{figure*}

\begin{table*}
\begin{center}
\begin{tabular}{rccccc}
\toprule
& $\mathcal{L}_D^R$ & $\mathcal{L}_D=\mathcal{L}_D^F$ & $\mathcal{L}_G$ & $\lambda$ & $\lambda^{-1}$\\
\midrule
Minimax \cite{goodfellow2014generative}             & $\splus(-y)$  & $\splus(y)$  & $-\splus( y)$ & $-1$       & $-1$ \\
Nonsaturating \cite{goodfellow2014generative}       & $\splus(-y)$  & $\splus(y)$  & $ \splus(-y)$ & $-e^{-y}$  & $-e^{y}$ \\
Wasserstein \cite{arjovsky2017wasserstein}          & $-y$          & $y$          & $-y$          & $-1$       & $-1$ \\
Least Squares \cite{mao2017least}                   & $(y-1)^2$     & $y^2$        & $(y-1)^2$     & $1-y^{-1}$ & $y\cdot(y-1)^{-1}$ \\
Hinge \cite{lim2017geometric, tran2017hierarchical} & $\relu(-y+1)$ & $\relu(y+1)$ & $-y$          & $\nexists$ & $-H(y+1)$ \\
\bottomrule
\end{tabular}
\end{center}
\caption{Common GAN losses and corresponding gradient scaling factors for FusedProp and InvFusedProp, where $\splus(x) = \ln(1+e^x)$, $\relu(x) = \max(x,0)$ and $H$ denotes the Heaviside step function.}
\label{table:losses}
\end{table*}

\section{Introduction}
Generative adversarial networks (GANs) have been continually progressing the state-of-the-art in generative modeling of all kinds of data since its invention \cite{goodfellow2014generative}.
Among its many applications, image generation arguably has received the most attention due to its strikingly realistic results \cite{karras2017progressive, karras2019style, karras2019analyzing, zhang2018self, brock2018large}.
However, the training of these powerful GANs usually takes days to weeks even on high-end multi-GPU/-TPU machines, strongly limiting the number of experiments researchers can afford and negatively affecting the fairness and progress of the field.

To mitigate this challenge, existing work mainly relied on two types of acceleration.
The first is to use lower numerical precision, \eg half precision (\texttt{fp16}) instead of single precision (\texttt{fp32}) for training \cite{karras2017progressive, karras2019style, karras2019analyzing, brock2018large}.
The second is to adapt GAN's architecture using \eg progressive growing \cite{karras2017progressive}, simplified normalization \cite{karras2019analyzing}, shared embedding \cite{perez2018film, brock2018large}, \etc

In this paper, we aim to accelerate the training procedure of GANs and propose the fused propagation (FusedProp) algorithm, a generalization of the gradient reversal algorithm \cite{ganin2015unsupervised} that can be used to train the discriminator and the generator of common GANs simultaneously using only one forward and one backward propagation.
Our algorithm offers $1.49\times$ the training speed compared to the conventional training of GANs and our code is publicly available.\footnote{\url{https://github.com/zplizzi/fusedprop}}
Although further studies are required to improve the stability of FusedProp, we hope our preliminary results and open-source implementation of FusedProp can accelerate future research on the training of GANs.

\section{Background}
The training of a GAN entails the minimax optimization of a two-player game between its discriminator $D$ and its generator $G$ defined as
\begin{equation}
\max_G\min_D\;\LD^R(D(x)) + \LD^F(D(G(z)))
\end{equation}
where $G$ is trained (by maximizing $\LD^F$) to map the latent variable $z$ from a given (\eg normal) distribution into $G(z)$ that resembles the real data $x$ such that $D$ can not tell $G(z)$ and $x$ apart even if it is trained (by minimizing $\LD^R$ and $\LD^F$) to do so.
It is rather common to write the optimization of $D$ and $G$ separately as
\begin{align}
\begin{split}
&\min_D\;\LD^R(D(x)) + \LD^F(D(G(z)))\\
&\min_G\;\LG(D(G(z)))
\end{split}
\end{align}
which allows for GAN losses with $\LG\neq-\LD^F$ and thus more desirable properties (\eg stronger gradients using the nonsaturating loss \cite{goodfellow2014generative}, see Table \ref{table:losses} and its references for more details).
For simplicity, we also write $\LD^F$ as $\LD$ in the rest of the paper.

\begin{align}
\begin{split}
\tD^{\,i+1} &= \tD^{\,i} - \alpha\frac{\pd\LD^R(D(x;\tD^{\,i})) + \LD(D(G(z;\tG^{\,i});\tD^{\,i}))}{\pd\,\tD^{\,i}}\\
\tG^{\,i+1} &=
\begin{cases}
\tG^{\,i} - \alpha\dfrac{\pd\LG(D(G(z;\tG^{\,i});\tD^{\,{\color{red}i  }}))}{\pd\,\tG^{\,i}} &\text{(SimGD)}\\
\tG^{\,i} - \alpha\dfrac{\pd\LG(D(G(z;\tG^{\,i});\tD^{\,{\color{red}i+1}}))}{\pd\,\tG^{\,i}} &\text{(AltGD)}
\end{cases}
\end{split}
\label{eq:simgd}
\end{align}

Although the training of $D$ and $G$ is often described as simultaneous, it is rarely the case in practice.
Specifically, instead of updating $\tD$ and $\tG$ simultaneously using SimGD \cite{mescheder2018training} as defined in Eq.~\eqref{eq:simgd},\footnote{Where $D(x)$ and $G(z)$ are written more precisely as $D(x;\tD)$ and $G(z;\tG)$ and stochastic gradient descent (SGD, instead of Adam) with learning rate $\alpha$ is used for simplicity.} updating them alternatingly using AltGD \cite{mescheder2018training} (often with multiple $\tD$ updates per $\tG$ update) is much more common, partly due to the stability and convergence concerns about SimGD \cite{salimans2016improved, mescheder2017numerics, mescheder2018training}.
However, researchers' view about SimGD is not unilaterally pessimistic since \cite{nagarajan2017gradient, heusel2017gans} proved SimGD can lead to stable convergence of GANs as well.
Encouraged by the positive results, we seek to accelerate the training of GANs based on the SimGD approach.

Of course, SimGD itself is not more computationally efficient than AltGD if one still needs to compute gradients for $\tD$ and $\tG$ using two backpropagations.\footnote{Which is equivalent to AltGD (\ie conventional) in Fig.~\ref{fig:fusedprop} except that the update for $\tD$ is delayed (till the update for $\tG$) and $z$ is reused (instead of redrawn for the second forward propagation).\label{footnote:simgd}}
Fortunately, it is known that if $\LG=\bar{\lambda}\LD$ for some constant $\bar{\lambda}$ (\eg $\bar{\lambda}=-1$ as in the minimax loss), the gradient reversal algorithm \cite{ganin2015unsupervised} originally designed for the domain adaptation problem can be used to combine the two backpropagations by inserting a simple function $\GR$ defined as
\begin{align}
\begin{split}
\GR_{\bar{\lambda}}(x) &= x \\
\frac{\pd\,\GR_{\bar{\lambda}}(x)}{\pd\,x} &= \bar{\lambda}I
\end{split}
\end{align}
between $D$ and $G$.\footnote{However, as also noted in \cite{tzeng2017adversarial}, using the gradient reversal algorithm with a common setting of $\bar{\lambda}=-1$ (\ie the minimax loss) to train GANs is not ideal \cite{goodfellow2014generative}, which may explain the lack of such attempts in the literature.}
Inspired by the gradient reversal algorithm, we aim to bring its level of efficiency to the training of GANs while supporting a broader set of GAN losses.

\section{Algorithm}
Although \cite{ganin2015unsupervised} also mentioned the possibility of generalizing the gradient reversal algorithm to arbitrary GAN losses, it is unclear if such generalization can be implemented as efficiently.
To this end, we formally derive the fused propagation (FusedProp) algorithm, a generalization of the gradient reversal algorithm for common GAN losses, and outline its implementation in the rest of the section.

The first form of FusedProp closely follows the gradient reversal algorithm, except with a data-dependent gradient scaling factor $\lambda$ for certain GAN losses.
As shown below
\begin{equation}
\frac{\pd\LG}{\pd\,\tG}
= \frac{\pd\LG}{\pd\LD} \frac{\pd\LD}{\pd\,G(z)} \frac{\pd\,G(z)}{\pd\,\tG}
= \frac{\pd\LD}{\pd\,G(z)} \underbrace{\frac{\pd\LG}{\pd\LD}}_{\lambda} \frac{\pd\,G(z)}{\pd\,\tG}
\end{equation}
and in Fig.~\ref{fig:fusedprop}, instead of computing $\frac{\pd\LG}{\pd\tG}$ with a second set of forward and backward propagations, one can\footnote{Due to the commutative property of the scalar and (Jacobian) matrix product.} scale $\frac{\pd\LD}{\pd G(z)}$ (a byproduct of computing $\frac{\pd\LD}{\pd\tD}$ during $\LD$ minimization) by $\lambda$ to extend the first backward propagation to obtain $\frac{\pd\LG}{\pd\tG}$, essentially fusing two sets of forward and backward propagations into one.
A PyTorch example of FusedProp training is provided in Fig.~\ref{fig:code1}.
For common GAN losses where $\LD$ and $\LG$ are both univariate scalar functions (\ie $\mathbb{R}\to\mathbb{R}$), $\lambda$ can be easily derived because $\frac{\pd\LG}{\pd\LD} = \frac{\LG'}{\LD'}$.
Table \ref{table:losses} summarizes $\lambda$ for 5 such GAN losses, where $\lambda$ is simply $-1$ for the minimax and the Wasserstein loss as in the gradient reversal algorithm, and depends on $y$ (the output of $D$) for the nonsaturating and the least squares loss.
The hinge loss however is not supported by this form of FusedProp, as the zero derivative part of $\relu$ leaves $\lambda$ undefined (division by zero).

\definecolor{codegreen}{rgb}{0,0.5,0}
\definecolor{codeblue} {rgb}{0,0,0.5}
\lstset{
  columns=fullflexible,
  basicstyle=\ttfamily\footnotesize,
  backgroundcolor=\color{white},
  commentstyle=\fontsize{8pt}{9pt}\color{codeblue},
  keywordstyle=\fontsize{8pt}{9pt}\color{codegreen},
  otherkeywords = {self},
  keepspaces=true,
  frame=tb,
  captionpos=b,
}
\begin{figure}
\begin{lstlisting}[language=python]
class FusedProp(torch.autograd.Function):
  @staticmethod
  def forward(ctx, Gz):
    return Gz

  @staticmethod
  def backward(ctx, gGz):
    # F1. get _lambda (calculated globally)
    global _lambda
    # F2. scale gGz (gradient of Gz) as Eq. (4)
    return gGz * _lambda.view(-1, 1, 1, 1)

# P1. forward D & G once with FusedProp
x_Gz = torch.cat(x, FusedProp.apply(G(z)))
yr, yf = D(x_Gz).chunk(2)
# P2. calculate loss (nonsaturating) and _lambda
loss_ns = F.softplus(-yr) + F.softplus(yf)
_lambda = -((-yf).exp())
# P3. backward D & G once to get all gradients
loss_ns.mean().backward()
# P4. update D & G simultaneously
optimizer_D_G.step()
\end{lstlisting}
\caption{PyTorch example of FusedProp training, where \texttt{P1-P4} describe the procedure of one training iteration and \texttt{F1-F2} describe the FusedProp steps.}
\label{fig:code1}
\end{figure}

\begin{figure}
\begin{lstlisting}[language=python]
class InvFusedPropLinear(torch.autograd.Function):
  @staticmethod
  def forward(ctx, x, W, b):
    ctx.save_for_backward(x, W)
    y = x.matmul(W.t()) + b
    return y

  @staticmethod
  def backward(ctx, gy):
    x, W = ctx.saved_tensors
    # I1. get _lambda_inv (calculated globally)
    global _lambda_inv
    # I2. pre-scale gy (gradient of y)
    scaled_gy = gy * _lambda_inv.view(-1, 1)
    # I3. Compute gradients for
    #     activation (x) with gy
    #     parameters (W & b) with scaled gy
    gx = gy.matmul(W)
    gW = scaled_gy.t().matmul(x)
    gb = scaled_gy.sum(0)
    return gx, gW, gb
\end{lstlisting}
\caption{PyTorch example of InvFusedProp-based linear (\ie fully connected) layer, where \texttt{I1-I3} describe the InvFusedProp steps. See our code for examples of other types of layers.}
\label{fig:code2}
\end{figure}

To circumvent the problem of the hinge loss, we propose the second form of FusedProp, the inverted FusedProp (InvFusedProp).
As shown below
\begin{equation}
\frac{\pd\LD}{\pd\,\tD}
= \frac{\pd\LD}{\pd\LG} \frac{\pd\LG}{\pd\,\tD}
= \frac{\pd\LG}{\pd\,\tD} \underbrace{\frac{\pd\LD}{\pd\LG}}_{\lambda^{-1}}
\end{equation}
and in Fig.~\ref{fig:fusedprop}, one can also obtain $\frac{\pd\LD}{\pd\tD}$ during $\LG$ minimization by scaling the ``incorrect'' gradient $\frac{\pd\LG}{\pd\tD}$ by $\lambda^{-1}$.
Worth to note, unlike FusedProp which can be trivially done in most deep learning frameworks, InvFusedProp requires additional effort to implement correctly and efficiently.\footnote{\Eg for convolutional layers, we need to use MKL-DNN or CuDNN subroutines for InvFusedProp to ensure performance.}
This is due to the fact that $\lambda^{-1}$ takes different values for different data in a batch, but in most frameworks gradients for parameters (here $\frac{\pd\LD}{\pd\tD}$) are only available as already reduced across all data in a batch for performance reasons.
Instead, one should pre-scale the gradient by $\lambda^{-1}$ before computing gradients for parameters within each layer of $D$.
A PyTorch example of InvFusedProp-based layer is provided in Fig.~\ref{fig:code2}.
InvFusedProp is slightly slower than FusedProp as additional scaling operations are needed in all layers of $D$.
For GAN losses with valid but different $\lambda$ and $\lambda^{-1}$ (\eg the nonsaturating and the least squares loss), it is also possible to adaptively switch between the two forms if the numerical accuracy of one is better than the other.\footnote{\Eg when using \texttt{fp16} for training. We do not observe such need using \texttt{fp32} in our experiments.}

Both forms of the FusedProp algorithm are exact and efficient implementations of the SimGD-based training of GANs, which bring the conventional time complexity of $\bigo(6\,T_D + 3\,T_G)$ down to $\bigo(4\,T_D + 2\,T_G)$, where $T_D$ and $T_G$ stand for the time complexities of the forward and backward propagations of $D$ and $G$ respectively.\footnote{This assumes $\LD^R$, $\LD$ and $\LG$ are all using the same batch size, and the gradients for parameters and activation within each layer are computed in parallel. If computed in serial, time complexities are $\bigo(8\,T_D + 4\,T_G)$ \vs $\bigo(6\,T_D + 3\,T_G)$.}
As $D$ and $G$ are commonly of similar complexity (\ie $T_D \approx T_G$), we can expect approximately $1.5\times$ theoretical speedup by using FusedProp training.
SimGD-based training of GANs however is not guaranteed to match the results of the conventional AltGD-based training, thus needs to be experimentally validated too.

\begin{table*}
\begin{center}
\begin{tabular}{ccccccccc}
\toprule
Architecture & $(D,G)$ LRs & Loss & Training & IS \cite{salimans2016improved} & FID \cite{heusel2017gans} & Speed\cref{footnote:speed} & Speedup & Samples \\
\midrule
\multirow{4}{*}{CNN}    & \multirow{4}{*}{$(2.0,2.0)\times10^{-4}$}
   & \multirow{2}{*}{NS} & C & $7.21\pm0.06$ & $27.23\pm0.96$ & $26.9$ \\
 & &                     & F & $7.17\pm0.05$ & $27.91\pm0.60$ & $41.7$ & $1.55\times$ & Fig.~\ref{fig:samples}.1\\
 & & \multirow{2}{*}{HG} & C & $7.32\pm0.12$ & $25.42\pm1.53$ & $27.0$ \\
 & &                     & I & $7.32\pm0.07$ & $24.59\pm1.13$ & $40.4$ & $1.50\times$ & Fig.~\ref{fig:samples}.2\\
\midrule
\multirow{4}{*}{ResNet} & \multirow{4}{*}{$(4.0,1.0)\times10^{-4}$}
   & \multirow{2}{*}{NS} & C & $7.66\pm0.20$ & $22.65\pm1.50$ & $14.9$ \\
 & &                     & F & $3.08\pm0.37$ & $118.2\pm14.7$ & $21.9$ & $1.47\times$ \\
 & & \multirow{2}{*}{HG} & C & $7.68\pm0.15$ & $19.93\pm1.66$ & $14.6$ \\
 & &                     & I & $4.11\pm0.39$ & $94.00\pm6.32$ & $21.1$ & $1.45\times$ \\
\midrule
\multirow{4}{*}{ResNet} & \multirow{4}{*}{$(4.0,0.5)\times10^{-4}$}
   & \multirow{2}{*}{NS} & C & $7.59\pm0.14$ & $27.56\pm1.17$ \\
 & &                     & F & $7.55\pm0.22$ & $26.97\pm1.67$ & & & Fig.~\ref{fig:samples}.3\\
 & & \multirow{2}{*}{HG} & C & $7.76\pm0.15$ & $23.41\pm1.15$ \\
 & &                     & I & $7.66\pm0.11$ & $23.49\pm0.79$ & & & Fig.~\ref{fig:samples}.4\\
\bottomrule
\end{tabular}
\end{center}
\caption{Unconditional CIFAR10 image generation results using conventional (C), FusedProp (F) or InvFusedProp (I) training, nonsaturating (NS) or hinge (HG) loss, and Adam optimizer at specified learning rates (LRs).}
\label{table:results}
\end{table*}

\begin{figure}
\begin{center}
\includegraphics[width=0.7\columnwidth]{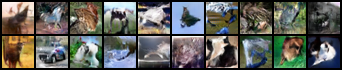}    \\[0.6em]
\includegraphics[width=0.7\columnwidth]{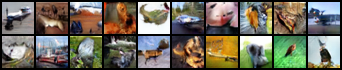}    \\[0.6em]
\includegraphics[width=0.7\columnwidth]{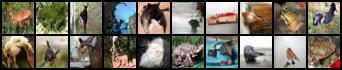} \\[0.6em]
\includegraphics[width=0.7\columnwidth]{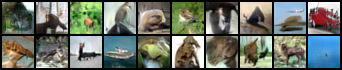}
\end{center}
\caption{Unconditional CIFAR10 image generation using Fused-Prop-trained GANs, where subfigures (numbered from top to bottom) come from experiments specified in Table \ref{table:results}.}
\label{fig:samples}
\end{figure}

\section{Experiments}
In this paper, we closely follow the setup of \cite{miyato2018spectral}, \ie unconditional CIFAR10 image generation using CNN or ResNet-based GANs with nonsaturating or hinge loss to validate the FusedProp algorithm.
We perform 5 runs for all configurations and summarize their Inception Scores (IS), Fréchet Inception Distances (FID) and speed\footnote{Measured in iterations per second at batch size of 64 for $\LD^R$, $\LD$ and $\LG$ using one V100 GPU.\label{footnote:speed}} in Table \ref{table:results}. 
Samples from the FusedProp-trained GANs are provided in Fig.~\ref{fig:samples}.

For CNN-based experiments, we choose the learning rate pair that performed the best in \cite{miyato2018spectral, kurach2018large} and find no significant difference in terms of IS and FID between conventional and FusedProp training.
For ResNet-based experiments, we first adopt the TTUR \cite{heusel2017gans} learning rate pair\footnote{Instead of multiple $\tD$ updates per $\tD$ update as suggested by \cite{kurach2018large} which we do not currently support.} used by \cite{zhang2018self} but find that FusedProp training performs significantly worse than conventional training in this setting.
With some manual tuning, we are able to stabilize FusedProp training and eliminate the difference in terms of IS and FID by halving the learning rate of $G$, which unfortunately also increases conventional training's FID, making this setting similar to \cite{kurach2018large} but likely worse than \cite{miyato2018spectral}.
On the other hand, we do observe sizable speedups using FusedProp training in all settings, ranging from $1.45\times$ to $1.55\times$ (overall $1.49\times$) which match the theoretical analysis.

Other factors that may cause a difference between conventional and FusedProp training are as follows.
First, conventional training implicitly uses twice the amount of power iterations in the spectral normalization compared to FusedProp.
Second, conventional training uses twice the amount of generated images in each iteration by redrawing $z$ compared to FusedProp.\cref{footnote:simgd}
However, we do not observe meaningful changes in the IS and FID when we correct conventional or FusedProp training to match each other in these two regards, implying that the fundamental difference between AltGD and SimGD-based training is the root cause here.\footnote{We have also tested SimGD without the FusedProp acceleration and obtained the same results as FusedProp, suggesting this is not due to any flaw in FusedProp.}

\section{Discussion}
Although our preliminary results indicate that FusedProp is not exactly a drop-in replacement for conventional training of GANs as it may require additional hyperparameter tuning due to SimGD's different nature, we hope that as more researchers start to realize and utilize its computational efficiency, more research will follow to fundamentally solve the issues of SimGD-based training.
At the same time, it will be crucial in our future work to study if existing techniques \cite{mescheder2018training, wiatrak2019stabilizing} can be efficiently combined with FusedProp to improve its stability for larger-scale problems.

The FusedProp algorithm also has known limitations, which we list as follows.
\begin{enumerate}[nolistsep]
\item
FusedProp does not provide much speedup if multiple $\tD$ updates are required per $\tG$ update \cite{arjovsky2017wasserstein, gulrajani2017improved}.
We find TTUR an effective replacement in our experiments and recommend using it instead, as also advocated by \cite{zhang2018self}.
\item
Gradient penalties on $D$ that involve $G(z)$, including \cite{gulrajani2017improved, kodali2017convergence} and the R2 penalty \cite{mescheder2018training}, are not compatible with FusedProp as their second-order gradients can incorrectly affect $G$.
The increasingly popular R1 penalty \cite{mescheder2018training, karras2019style, karras2019analyzing} however is compatible.
\item
Most conditional GANs \cite{mirza2014conditional, reed2016generative, miyato2018cgans} are compatible with FusedProp.
However, ones that explicitly use a classification loss in addition to the GAN loss \cite{odena2017conditional} are not compatible as gradients from those two losses become inseparable to be correctly scaled.
\end{enumerate}

{\small
\bibliographystyle{ieee_fullname}
\bibliography{main}
}

\end{document}